%% file: a_paper.tex
\documentclass{bmvc2k}

\usepackage{comment}

\title{Revisiting Self-Supervised Contrastive Learning for Facial Expression Recognition}

\addauthor{Yuxuan Shu}{y.shu21@imperial.ac.uk}{1}
\addauthor{Xiao Gu}{xiao.gu17@imperial.ac.uk}{1}
\addauthor{Guang-Zhong Yang}{gzyang@sjtu.edu.cn}{2}
\addauthor{Benny Lo}{benny.lo@imperial.ac.uk}{1}

\addinstitution{
 The Hamlyn Centre\\
 Imperial College London\\
 London SW7 2AZ, UK
}
\addinstitution{
 The Institute of Medical Robotics \\
 Shanghai Jiao Tong University\\
 Shanghai 200240, China
}

\runninghead{SHU ET AL.}{Revisiting Self-Supervised Contrastive Learning for FER}

\def\etal{\emph{et al}\bmvaOneDot}

\usepackage{times}
\usepackage{epsfig}
\usepackage{graphicx}
\usepackage{amsmath}
\usepackage{amssymb}
\usepackage{multirow}
\usepackage{enumerate}
\usepackage{colortbl}
\usepackage{tabularx}
\definecolor{maroon}{cmyk}{0,0.87,0.68,0.32}
\usepackage{algorithm}
\usepackage{algpseudocode}
\usepackage{booktabs}
\usepackage{adjustbox}
\usepackage{siunitx}
\usepackage{dsfont}
\usepackage{comment}
\usepackage{enumitem}
\setlist{noitemsep,topsep=0pt,parsep=0pt,partopsep=0pt}

\newcommand{\bx}{\boldsymbol{x}}

\usepackage{color}
\definecolor{my_color1}{rgb}{0.94,0.87,0.8}
\definecolor{my_color2}{rgb}{0.63,0.79,0.95}
\definecolor{my_color3}{rgb}{1.0,0.79,0.95}
\definecolor{color_red}{RGB}{233,36,79}
\definecolor{color_yellow}{RGB}{249, 209, 60}
\definecolor{color_green}{RGB}{34, 139, 34}
\definecolor{Line}{rgb}{.5,.5,1}

\begin{document}

\maketitle

\begin{abstract}

The success of most advanced facial expression recognition works relies heavily on large-scale annotated datasets. However, it poses great challenges in acquiring clean and consistent annotations for facial expression datasets. On the other hand, self-supervised contrastive learning has gained great popularity due to its simple yet effective instance discrimination training strategy, which can potentially circumvent the annotation issue. Nevertheless, there remain inherent disadvantages of instance-level discrimination, which are even more challenging when faced with complicated facial representations. In this paper, we revisit the use of self-supervised contrastive learning and explore three core strategies to enforce expression-specific representations and to minimize the interference from other facial attributes, such as identity and face styling. Experimental results show that our proposed method outperforms the current state-of-the-art self-supervised learning methods, in terms of both categorical and dimensional facial expression recognition tasks. Our project page: \url{https://claudiashu.github.io/SSLFER}.

\end{abstract}

\vspace{-10pt}
\section{Introduction}
\vspace{-5pt}
\input{section/intro}

\vspace{-5pt}
\section{Related Works}
\vspace{-5pt}
\input{section/related}

\section{Methodology}
\vspace{-5pt}
\input{section/method}

\section{Experiments}
\vspace{-5pt}
\input{section/experiment}

\vspace{-10pt}
\section{Discussions and Conclusions}
\vspace{-5pt}
\input{section/discussion}

\clearpage

\bibliography{egbib}

\clearpage

\appendix

\input{section/appendix}

\end{document}

%% file: section/intro.tex
\input{image/first}

Facial expression recognition (FER) plays an important role in a series of applications, ranging from human-computer interaction~\cite{dix2004human}, social robotics~\cite{breazeal2016social}, to mental health monitoring~\cite{fei2020deep}. Recently, considerable research efforts in computer vision have been dedicated to developing systems that can understand facial expressions from facial images automatically, including basic emotion categories in the categorical level~\cite{ekman1971constants}, as well as valence and arousal in the dimensional level~\cite{Baron2003texbook}. 

One of the reasons leading to the success of most FER systems is the availability of large-scale annotated facial datasets~\cite{mollahosseini2017affectnet,kollias2022abaw}. However, the curation of such datasets poses several challenges over the course of acquiring labels. In fact, annotating real-world facial images requires significant amount of time/efforts and high-level expertise~\cite{kollias2022abaw}. Even worse, different levels of expertise may induce annotation bias, leading to inconsistent or noisy labels~\cite{zeng2018facial}, and the category distribution is usually imbalanced~\cite{gu2022tackling} such that the performance of direct supervised learning is limited. Faced with these issues, there is a pressing need for minimizing the reliance of annotated data, yet achieving satisfactory FER performance, and self-supervised learning can potentially be the solution.

Self-supervised learning (SSL) is an emerging research line in deep learning, which aims to derive semantically meaningful representations by self-designed proxy tasks~\cite{liu2021self}. Among existing solutions, contrastive based SSL has demonstrated reasonable performance by instance-level discrimination~\cite{chen2020simple,he2020momentum,grill2020bootstrap}. It functions by maximizing the similarity, in the representation level, of the same image under different augmentations/views (positive pair), whereas minimizing those of different instances (negative pair). Despite its success on several vision tasks, there remain inherent issues of such instance discrimination strategy~\cite{zhang2022towards,robinson2020contrastive}, which become even more apparent when being applied on FER. 

As shown in \figureautorefname~\ref{fig:ins}, 
given a batch of samples drawn from different subjects, what would be learned when promoting the similarity of images under different augmentations? In fact, facial images are complicated, composed of multiple factors/attributes including expressions, head poses, identity, makeup, hair styles, etc.~\cite{chang2021learning}. It is hardly possible to regulate the network to learn expression-relevant features only, without knowing the actual expression label. We hereby revisit self-supervised contrastive learning, and argue that there are three main promising solutions for better facial expression-related representation learning when performing instance discrimination.

\textbf{1. Exploring More Effective Augmentations for Positive Pairs.} Stronger augmentations have been proven to be able to facilitate better representation learning~\cite{li2022contrast}. Conventional spatial augmentation operations utilized in exiting visual image recognition works~\cite{chen2020simple,he2020momentum}, like crop/resize and horizontal flipping, may not be sufficient for extracting robust expression related information. 

\textbf{2. Increase Hard Negative Pairs.}
Hard negative pairs represent images with different downstream labels, yet not easy to be differentiated~\cite{robinson2020contrastive}. Increasing the number of hard negative pairs during training is able to force the network to learn task related features. However, without knowing the actual emotion label in self-supervised settings, it is difficult to determine \textcolor{black}{which is hard} and which is negative. It remains open to develop an effective strategy to pick up the real hard negative pairs to facilitate training.

\textbf{3. Reduce False Negative Pairs.}
Furthermore, given a large batch of data, there would be quite a few samples having similar expressions, yet treated as negative pairs to be pushed away. The existence of such false negative pairs would further limit the performance of self-supervised contrastive learning~\cite{huynh2022boosting}. However, it is challenging to reduce false negative pairs without knowing the actual labels.

Therefore in this paper, in line with these three core directions, we propose an effective self-supervised contrastive learning framework for FER. Based on expression related spatio-temporal augmentations and the region-level similarity in similar expressions, we complementary intergrate several novel strategies to mitigate the above three issues in FER. The proposed method was evaluated on two different FER tasks. 

%% file: image/first.tex
\begin{figure}[h]
\begin{center}
    \setlength{\tabcolsep}{3pt}
    \resizebox{\linewidth}{!}{
    \begin{tabular}{cccc}
    \includegraphics[width=0.3\linewidth]{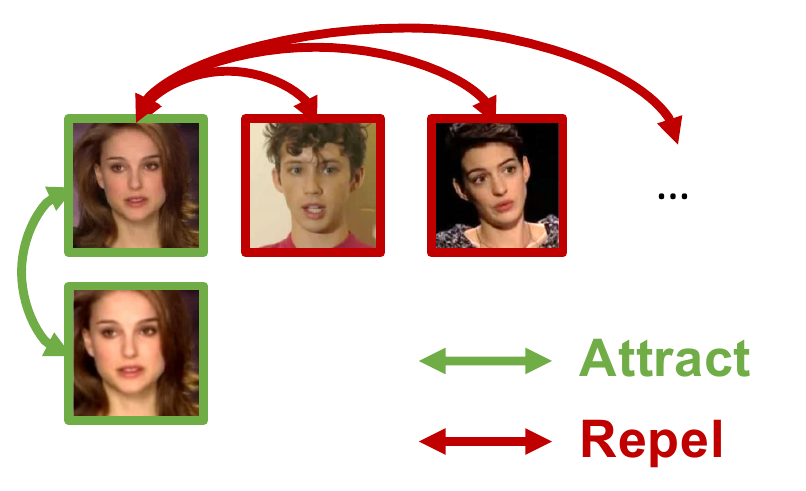}\label{fig:ins}&
    \includegraphics[width=0.2\linewidth]{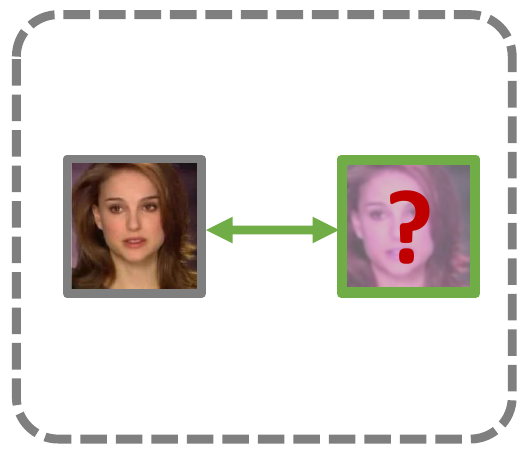}\label{fig:aug}&
    \includegraphics[width=0.2\linewidth]{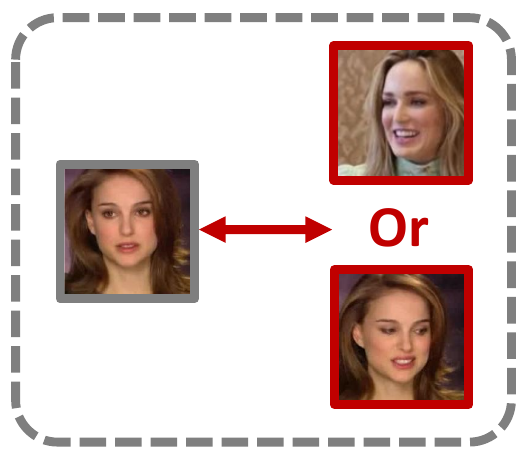}\label{fig:hn}&
    \includegraphics[width=0.2\linewidth]{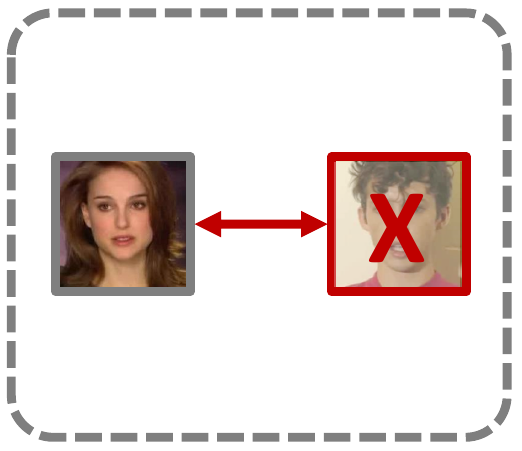}\label{fig:fn}\\
    (a) \small{Instance Discrimination} & (b) \small Augmentations & (c) \small {Hard Negatives} & (d) \small {False Negatives}
    \end{tabular}}
    \vspace{-10pt}
    \caption{\textbf{Illustration of the inherent issue and potential solutions when applying self-supervised instance-level contrastive learning for facial expression recognition.} (a) The conventional paradigm for self-supervised contrastive learning aims to pull close image representation from the same instance together, where emotion irrelevant features, e.g human identities and hair styles, would be easily learned as shortcuts. In this paper, we propose three core directions to tackle this. (b) \textbf{Explore effective data augmentations}, Section~\ref{sec:aug}. (c) \textbf{Increase hard negative pairs}, Section~\ref{sec:hn}. (d) \textbf{Reduce false negative pairs}, Section~\ref{sec:fn}. Pictures are selected from VoxCeleb~\cite{nagrani2017voxceleb,chung2018voxceleb2,nagrani2020voxceleb}.}
\end{center}
\vspace{-15pt}
\label{fig:first}
\end{figure}

%% file: section/related.tex
\textbf{Facial Expression Learning.} Thus far, considerable research efforts have been devoted to facial expression recognition, in terms of advanced computational models that are able to extract discriminative FER features~\cite{farzaneh2021facial,zhao2018feature,xue2021transfer}. Despite these advances, most of these solutions rely heavily on annotated training data, which would fail when faced with insufficient high-quality and consistent labels. \textcolor{black}{A few recent works have proposed the use of self-supervised learning for expression~\cite{roy2021self,khare2021self}. They are heavily dependent on specific datasets, whereas multi-view images~\cite{roy2021self} or additional modalities~\cite{khare2021self} are required. It remains an open research question on how to design effective SSL tasks, to effectively extract expression-specific information.}

\noindent\textbf{Facial Attribute Decomposition.} Another line of research is targeted at decomposing different attributes from facial representations. This is important for learning robust expression-related features, since it reduces the spurious correlation caused by other irrelevant attributes (such as styling and identity). Existing works have attempted to perform disentanglement of identity and facial expressions~\cite{wu2020cross}, which however mostly are conducted in a supervised manner. On the other hand, recent studies have investigated unsupervised ``Deepfake'' (face swapping) approaches~\cite{rossler2019faceforensics++,Li2019faceshifter,perov2020deepfacelab}. One representative work \cite{chang2021learning} proposed a cycle-consistent framework to disentangle ``expression'' and identity apart. However, since facial embeddings are complicated, the authors also indicated that other attributes, such as head poses, could also be learned into the expression representations. This degrades the generalization of FER, and as reported in \cite{chang2021learning}, the proposed self-supervised training is still inferior to pure fully-supervised training for FER by a large margin.

\noindent\textbf{Self-Supervised Contrastive Learning.}
In recent studies, self-supervised learning has been proven to have a great potential in learning good visual representations~\cite{jaiswal2020survey} by either generative or contrastive, or the combinations~\cite{liu2021self}. Contrastive learning is emerging as a simple yet effective self-supervised manner, based on instance discrimination~\cite{chen2020simple,he2020momentum,grill2020bootstrap,qian2021spatiotemporal}. However, its success relies on the assumption that the training batch are i.i.d (independently and identically distributed) sampled~\cite{zhang2022towards}, which however is rarely the case for facial expression datasets~\cite{zhang2022towards}, since the facial expressions are complicated with many other irrelevant factors, such as head poses and identity-related features. It is of paramount importance to avoid the shortcuts caused by these irrelevant factors, yet difficult without knowing the labels.

Approaches could be focused on either the instances that are going to be attracted (related to data augmentations~\cite{qian2021spatiotemporal}), or to be repelled (related to hard pair selection~\cite{huynh2022boosting,robinson2020contrastive}); however; these have not been systematically explored in the challenging FER task yet. This leads to the questions that we want to discuss: 
\begin{itemize}
    \item In terms of image augmentation, what information should be preserved or discarded to boost the facial-expression representation learning.
    \item Is there any information that could act as a supervisory signal to help enhance the model performance in FER related contrastive learning.
\end{itemize}

%% file: section/method.tex
\subsection{Notations and Problem Definition}

Contrastive learning works by clustering the positive pairs (usually generated with different augmentations on the same image or aligned modality), whereas pushing away the negatives (usually generated from other images). Given a batch of data $\boldsymbol{X}=\{\bx_{1}, \bx_{2}, ... \bx_{N}\}$, the \textit{InfoNCE Loss} for self-supervised instance discrimination is formulated as in \equationautorefname~ \eqref{eq:infonce}.

\begin{equation}
\footnotesize
\mathcal{L}_{N}=\mathbb{E}_{\boldsymbol{X}}\left[-\log \frac{e^{f_{sim}\left(\bx_{i}, \bx_{j}\right)/\tau}}{\sum_{k=1}^{N} \mathds{1}_{[k \neq i,j]} e^{f_{sim}\left(\bx_{i}, \bx_{k}\right)/\tau}}\right],
\label{eq:infonce}
\end{equation}

 where {$N$} is the number of images in a batch,  $\tau$ is a temperature constant, and {$f_{sim}$} is the similarity matrix of feature representations, which is cosine similarity by default. For each anchor image \cal{$\bx_{i}$}, the image \cal{$\bx_{j}$} (different augmentations of $\bx_{i}$), is considered as positive. The objective of this loss is to minimize the distance between positive pairs, while maximizing that between other negative image pairs.
 
 However, for self-supervised FER, it is challenging to regulate the network to only learn expression-related representations without the interference of other attributes, if no prior knowledge is given. To tackle this, we present our solutions as below. 

\vspace{-5pt}\subsection{Positives with Same Expression}\vspace{-5pt}
\label{sec:aug}

\begin{figure}[ht]
\vspace{-15pt}
\centering
  \begin{minipage}[c]{0.47\linewidth}
  \centering
    \includegraphics[width=0.8\linewidth]{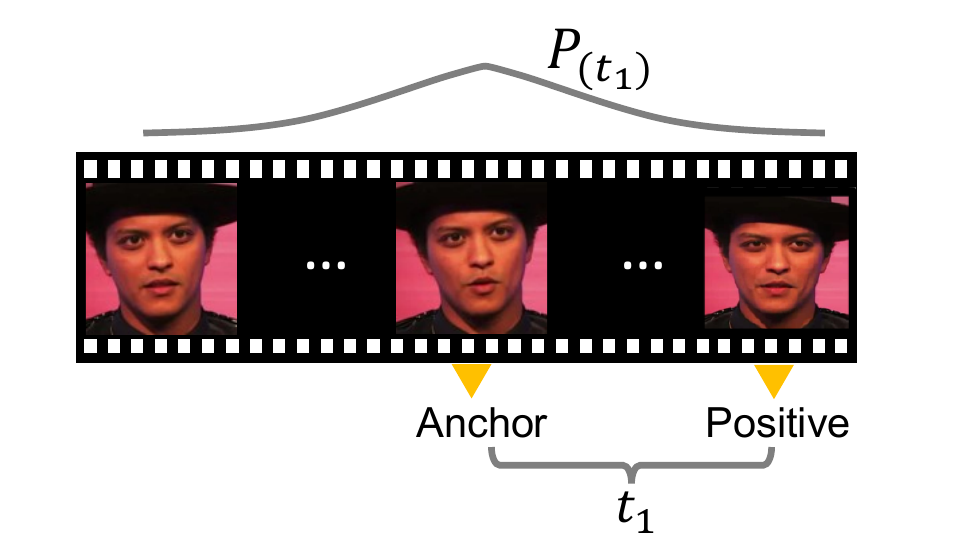}
    \vspace{-10pt}
    \caption{\textbf{TimeAug:} We sample the video sequence (Voxceleb1~\cite{nagrani2017voxceleb}) along the time domain to perform time augmentation. } \label{fig:timeAug}
  \end{minipage}\hfill
  \begin{minipage}[c]{0.48\linewidth}
    \centering
    \includegraphics[width=0.8\linewidth]{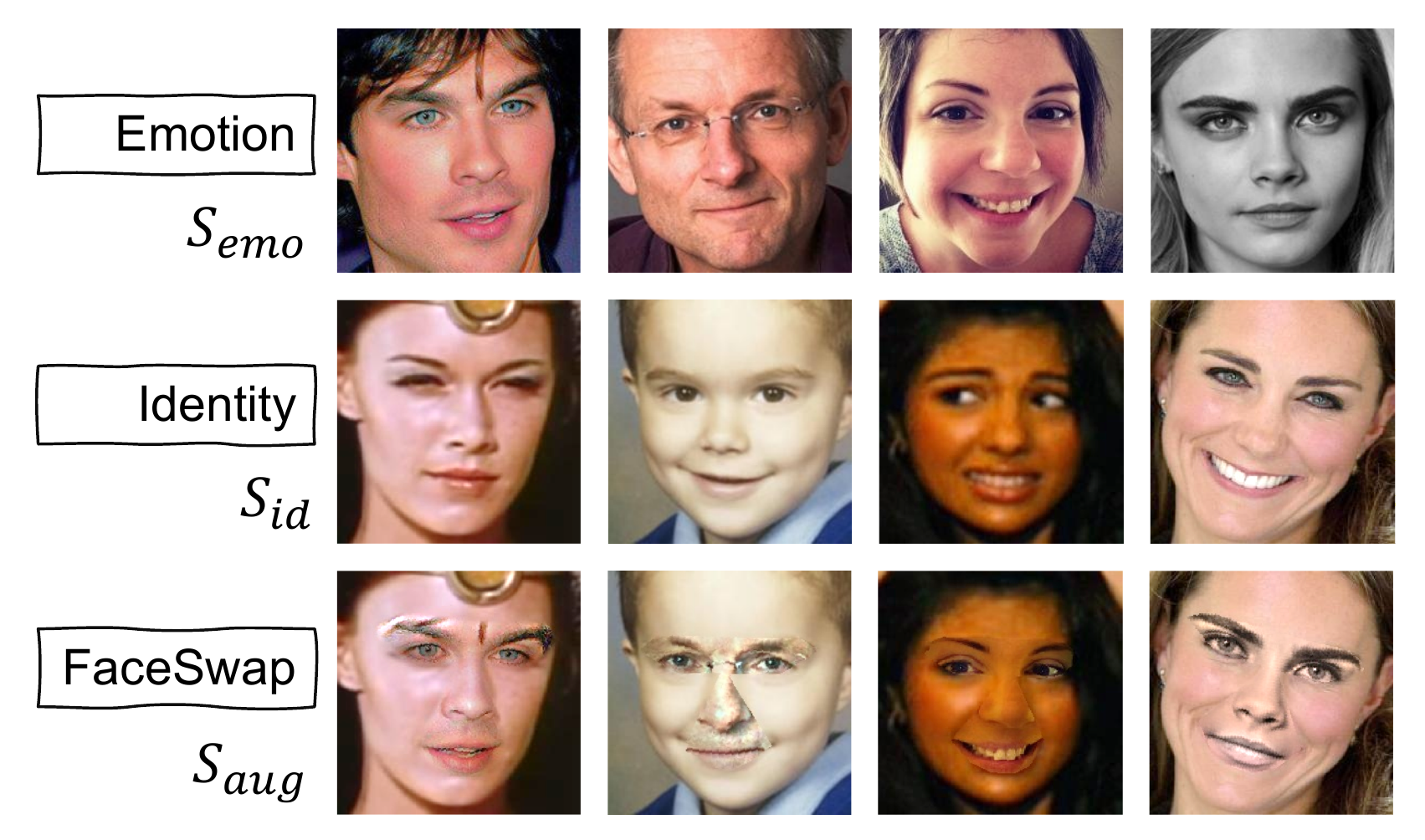}
    \vspace{-10pt}
    \caption{\textbf{FaceSwap:} {Emotion is transferred to another face with different identities to synthesize positive pairs. (Pictures selected from AffectNet~\cite{mollahosseini2017affectnet} for  demonstration)}} \label{fig:faceswap}
  \end{minipage}
\vspace{-10pt}
\end{figure}

First of all, to explore effective augmentations, we developed stronger augmentation strategies mainly based on two assumptions: 1) Expressions of human beings tend to change slightly within a short time interval.
2) The facial appearance indicates more identity-related information compared to facial landmark structural characteristics.

\vspace{-5pt}
\subsubsection{Temporal Augmentation}\vspace{-5pt}
\label{sec:tempaug}

We adopt temporal augmentation (\textbf{TimeAug}) by adding temporal shifts to positive pair generation. It is an intuitive concept that two samples sampled from the same video possess higher similarity with smaller time intervals. Based on this observation, we assume that within a short time period, facial expressions would not vary too much, and the shorter the time interval, the more similar their expressions would be. In practice, inspired by~\cite{qian2021spatiotemporal}, the sampled time interval $t_1$ follows a downscale distribution over $[0, T_1]$ (\figureautorefname~\ref{fig:timeAug}), where $T_1$ represents the maximum time interval that is considered as positives (empirically 1 second) and $P({t}_{1})$ stands for the probability of sampling ${t}_{1}$.

\vspace{-5pt}
\subsubsection{FaceSwap}\vspace{-5pt}

\label{sec:swap}

Whilst adding \textbf{TimeAug} to the training procedure enhances the capability of extracting expression-related information, it remains unresolved how to pull close the images of different persons which are supposed to act as positives (i.e. the same expression). To facilitate this, we resort to a simple yet effective strategy \textbf{FaceSwap} to generate fake faces that exhibits a similar expression as shown in \figureautorefname~\ref{fig:faceswap}, built on the assumption that facial appearance represents more identity-related information rather than expressions compared to facial landmark structures. The procedure is illustrated as below:
\begin{enumerate}
    \item Given a positive sample ${S}_{emo}$ (provides the same expression as the Anchor image) and an image ${S}_{id}$ that is randomly selected from the whole training set (provides a different identity from the Anchor image), extract the landmarks of the two images.
    \item Align the two images with colour correction and topology transformation according to the facial landmarks.
    \item Replace the region within the landmark in ${S}_{id}$ with ${S}_{emo}$.
\end{enumerate}

To the best of our knowledge, this is the first work that adopts face swapping strategy into contrastive learning as an augmentation strategy. It should be noted that this procedure is similar to another line of facial works, ``Deepfake''~\cite{rossler2019faceforensics++,Li2019faceshifter,perov2020deepfacelab}. Instead of utilizing those deep-learning based face-swapping methods, our strategy has already demonstrated its efficacy and is far easier and more computationally efficient. In practice, we performed on 50\% of the positive images, because our proposed simple operation might generate weird faces when there are significant facial differences, and structural information cannot fully represent facial expressions.

\vspace{-12pt}\subsection{Negatives with Same Identity}\vspace{-5pt}
\label{sec:hn}

\label{section:hardNeg}

To further avoid the shortcuts caused by identity-related information during instance discrimination, we propose \textbf{HardNeg}. On top of the original negative pairs from other identities, we sample images from the same subject but with large time interval as ``hard negative''. Given two facial images from the same identity, the representation will least prefer identity-related information if we are going to push away them. \textcolor{
black}{Hard negative pairs with $t_2$ interval are randomly sampled from $T_2$ onwards, where $T_2$ is the minimum time interval that is considered as hard negative (empirically as 3 seconds).}

\begin{figure}[ht]
\vspace{-5pt}
\begin{center}
\includegraphics[width=0.75\linewidth]{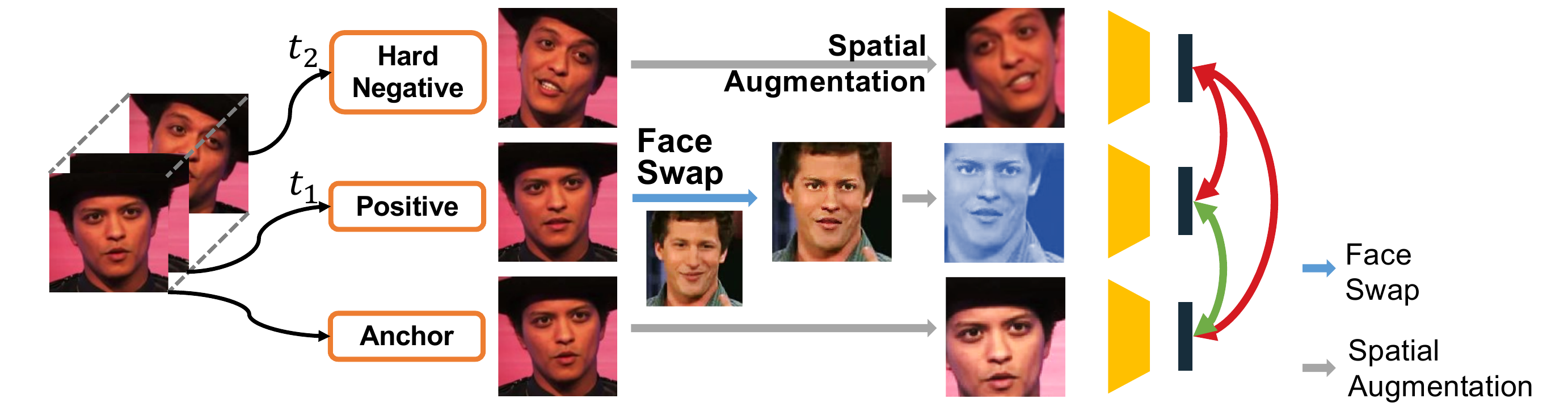}
\end{center}
\vspace{-15pt}
  \caption{\textbf{Illustration of the positive and hard negative pair generation.} Positive images are chosen with random \textbf{TimeAug} and augmented randomly with \textbf{FaceSwap}. Meanwhile, hard negative pairs are selected from the images with the same identity but large time intervals to avoid identity-related information to be learned. (Demonstrate with Voxceleb1~\cite{nagrani2017voxceleb}.)} \label{fig:structure}
\vspace{-10pt}
\end{figure}

Thus far, the whole training pipeline is shown in \figureautorefname~\ref{fig:structure}, with both effective augmentations for positive pair generation and sampling policy for hard negative pair selection being incorporated into. However, the negative, including the hard negatives we select, could potentially be false negatives. The urgent issue on false negatives is still unresolved.

\vspace{-10pt}\subsection{False Negatives Cancellation}\vspace{-5pt}
\label{sec:fn}

\begin{figure}[tp!]

\begin{center}
\includegraphics[width=0.75\linewidth]{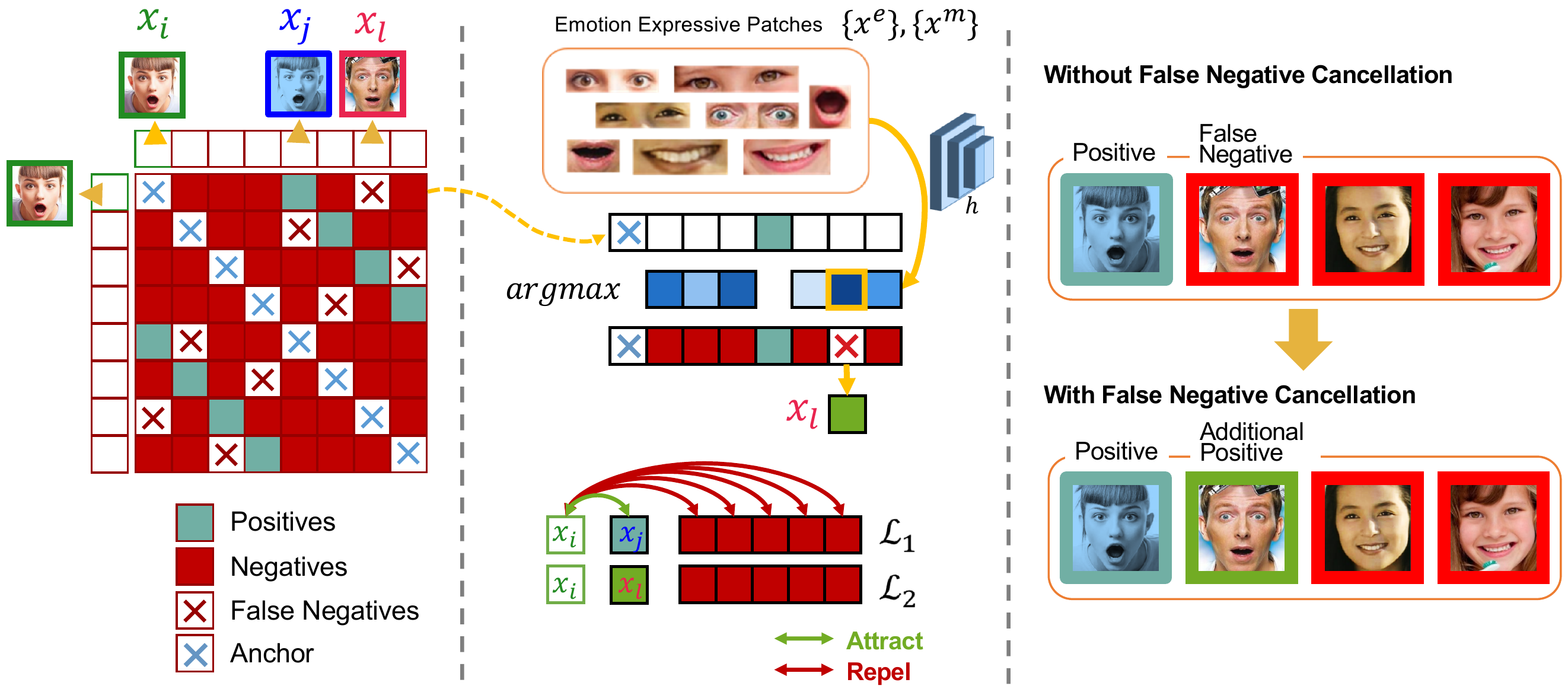}
\end{center}
\vspace{-15pt}
  \caption{\textbf{Illustration of false negative cancellation.} {Eyes and mouths are cropped as expressive indicators, with features being extracted and concatenated for similarity matrix calculation. We pick the sample with the highest similarity to the anchor, considering it as a false negative. Subsequently, we calculate the loss based on the original positive target \(\mathcal{L}_{1}\) as well as the picked false negative \(\mathcal{L}_{2}\).} Pictures are selected from AffectNet~\cite{mollahosseini2017affectnet}. }

\label{fig:mask_structure}
\vspace{-15pt}
\end{figure}

In FER, false negatives are the images that are treated as negatives, yet they actually present similar expressions to the anchor image. False negatives are undesirable since they impair the training process by discarding the semantic information of the same expression, especially when there are only a small number of classes when training with a large batch size. Unfortunately, this is difficult to avoid without knowing the actual labels. Thus we propose \textbf{MaskFN} to help minimizing the negative effects caused by false negatives during training, as shown in \figureautorefname~\ref{fig:mask_structure}. Here we consider the facial expressions as discrete emotions~\cite{ekman1971constants}.

In fact, some landmark areas, like eyes and mouth, usually contain more explicit expression signs than the other facial areas. For instance, the emotion ``Happy'' usually contains a smile with the corners of the mouth turning up, while ``Fear'' usually comes with wild open eyes and mouth. Robert and Adam~\cite{robert2016mouth} studied the low-level mouth feature-based emotion classification, and have proved that mouth could act as an expressive indicator of emotion. 

On the basis of the above observation, we assume that:
\begin{itemize}
    \item High-level features of eyes and mouths have higher similarity for those facial images with similar emotions. 
    \item Those recognized as false negatives can be set as positives of the anchor image.
\end{itemize}

Given the assumptions and the fact that the batch size is much larger than the class numbers, in a mini-batch, the pair with most similar eye or mouth regions are of high probability to be false negatives. Therefore, we propose a novel strategy to identify the potential false negative by utilizing the characteristics of eyes and mouth, implemented as follows,

\begin{enumerate}
    \item For each image \cal{$\bx$} in \cal{$\boldsymbol{X}$}, extract the landmarks of the face, and crop the region of eyes \cal{$x^{e}$} and mouth \cal{$x^{m}$} accordingly.
    \item Obtain \cal{$z^e, z^m=h(x^{e}), h(x^{m})$}, where \cal{$h$} is the feature encoder (\textit{ResNet18 pretrained on ImageNet}) and \cal{$z^{e}, z^{m}$} are the globally averaged features.
    
    \item Concatenate \cal{$z^{e}, z^{m}$} into \cal{$z^{cat}$} and calculate \cal{$f_{sim}$} of feature \cal{$z^{cat}$} with cosine similarity. 
    \item Set \cal{$N_{FN} (N_{FN}<<N)$} images with the highest similarity (that is originally neither positive nor hard negative) as additional positives.
\end{enumerate}

The proposed false negative cancellation pipeline is shown in \figureautorefname~\ref{fig:mask_structure}. Without loss of generality and for clarity, we only discuss picking up one false negative in \textbf{MaskFN} for each anchor by default in this paper, if not specified. The contrastive loss is calculated as shown in \equationautorefname~\eqref{eq:rep}, where \(\bx_{l}\) stands for the additional positive image with respect to anchor image \(\bx_{i}\). \(\mathcal{L}_{2}\) is assigned with less weight (empirically set as $\frac{1}{2}$), since the selected false negative images may not be the true positive pairs as we only consider the similarity of mouth and eyes as criteria. The total loss is formulated as $\mathcal{L}_{1}+\frac{1}{2}\mathcal{L}_{2}$. 

\begin{equation}
\vspace{-10pt}
\footnotesize
  \mathcal{L}_{1} = \mathbb{E}_{X}  \left[-\log \frac{e^{f_{sim}\left(\textcolor{color_green}{\bx_{i}},\textcolor{blue}{\bx_{j}}\right)/\tau}}{\sum_{k=1}^{N} \mathds{1}_{[k \neq \textcolor{color_green}{i},\textcolor{blue}{j},\textcolor{color_red}{l}]} e^{f_{sim}\left(\textcolor{color_green}{\bx_{i}}, \bx_{k}\right)/\tau}}\right], \;\;\;\;\;\;
  \mathcal{L}_{2} = \mathbb{E}_{X}  \left[-\log \frac{e^{f_{sim}\left(\textcolor{color_green}{\bx_{i}}, \textcolor{color_red}{\bx_{l}}\right)/\tau}}{\sum_{k=1}^{N} \mathds{1}_{[k \neq \textcolor{color_green}{i},\textcolor{blue}{j},\textcolor{color_red}{l}]} e^{f_{sim}\left(\textcolor{color_green}{\bx_{i}}, \bx_{k}\right)/\tau}}\right].
\label{eq:rep}
\end{equation}

%% file: section/experiment.tex
\subsection{Experiment Setup}\vspace{-5pt}

\noindent\textbf{Self-supervised pre-training.} All self-supervised learning network was trained on NVIDIA GeForce RTX 3090 from scratch. ResNet50 was used as the backbone network for feature extraction. Following SimCLR \cite{chen2020simple}, two MLP (Multi-Layer-Perceptron) layers were added with a default 128 feature dimension. The training batch size was 256 with a initialised 3e-4 learning rate and 1e-4 weight decay. \textcolor{black}{We optimized the network using Adam optimizer except for MoCo-v2~\cite{chen2020improved}, BYOL~\cite{grill2020bootstrap} optimized with SGD.} We applied a cosine annealing schedule from 10 epochs onwards. The temperature $\tau$ was set to 0.07 in the InfoNCE loss for all experiments. \textcolor{black}{During this stage, we saved the checkpoints every 5 epochs when its instance discrimination top-1 accuracy reached 60\% for downstream task evaluation.}

\noindent\textbf{Downstream evaluation.} We evaluated the pretraining performance by freezing and fine-tuning the encoder with three MLP layers on top. For fair comparison, we kept the same set of hyper-parameters across all models for each downstream task, with a batch size of 64. \textcolor{black}{Training was conducted using an Adam optimizer over 20 iterations for AffectNet dataset and 300 iterations for FER2013 dataset with a weight decay of 5e-4. Initially, the learning rate was set to 1e-4 and decayed with cosine annealing learning rate scheduler.}

\noindent\textbf{Data augmentation.} 
The adopted spatial augmentation for both contrastive learning and downstream evaluation include landmark-based masking (eyes or mouth), resizing (128, 128), random crop (112, 112), horizontal flipping, Gaussian blurring, colour jittering, and random grayscale. \textcolor{black}{We applied \textbf{TimeAug}, \textbf{FaceSwap} only in self-supervised pre-training.}

\noindent\textbf{Landmarks.} To facilitate \textbf{FaceSwap} in Section \ref{sec:swap}, we used dlib toolkit \cite{king2009dlib} to detect landmarks and the bounding box of eyes and mouth. The extracted landmark contains 68 points, including jaw, left eyebrow, right eyebrow, left eye, right eye, nose and mouth.

\vspace{-12pt}\subsection{Datasets}\vspace{-5pt}

\textcolor{black}{Three} datasets were used in our work, namely \textbf{VoxCeleb1}, \textbf{AffectNet} and \textbf{FER2013}.

The \textbf{VoxCeleb1}~\cite{nagrani2017voxceleb} dataset is a large-scale facial dataset, which includes videos that were extracted from YouTube of 1251 celebrities with different ages and of different ethnicity. It provides image sequences yet no emotion related annotations.  

The \textbf{AffectNet}~\cite{mollahosseini2017affectnet} dataset is the largest facial expression annotated (classification and regression) in the wild dataset which contains more than 1,000,000 facial images from different genders and races. We used all 8 classes of emotions (neutral, happy, angry, sad, fear, surprise, disgust, contempt) for classification on the publicly available training and validation splits, with the same settings as in~\cite{wang2020region}.

The \textbf{FER2013}~\cite{fer2013challenge} dataset is a wildly used dataset in grayscale which consists of 28,709 training images. We reported the results on the test split, with the best performance model on the validation split, \textcolor{black}{following the same settings as~\cite{minaee2021deep,vulpe2021convolutional}}.

The proposed method was trained on the \textbf{VoxCeleb1}~\cite{nagrani2017voxceleb}. After training, the performance of the learnt representation was validated on \textbf{AffectNet}~\cite{mollahosseini2017affectnet} and \textbf{FER2013}~\cite{fer2013challenge} with two downstream tasks, namely Emotion Classification and Valence \& Arousal Recognition.

\vspace{-5pt}\subsection{Downstream Tasks}\vspace{-5pt}

\noindent\textbf{Emotion Classification:} Macro ${F}_{1}$ score, ${F}_{1} = \frac{1}{n} \sum_{i=1}^{n}\mathcal{F}_{1}^{C_i}$, was used for evaluating categorical expression classification. Since the training set may suffer from imbalance, we adopted the state-of-the-art {Balanced Softmax Cross Entropy Loss}~\cite{ren2020balanced} for training.  

\noindent\textbf{Valence \& Arousal Recognition:}
We also reported results of Valence \& Arousal recognition, where two common metrics were used for evaluating both the error and correlation. They are Root Mean Square Error $RMSE=\sqrt{\frac{1}{n}\sum_{i=1}^{n}(\hat{\theta}_i-\theta_i)^2}$ and Concordance Correlation Coefficient (CCC)~\cite{lawrence1989concordance}  $\rho_{c}=\frac{2 \rho \sigma_{\hat{\theta}} \sigma_{\theta}}{\sigma_{\hat{\theta}}^{2}+\sigma_{\theta}^{2}+\left(\mu_{\hat{\theta}}-\mu_{\theta}\right)^{2}}$, respectively. We applied CCC loss for regression optimization following~\cite{meng2022valence}.

\vspace{-10pt}\subsection{Quantitative Results}
\vspace{-5pt}

\input{table/table_finetune}

In this section, quantitative results are presented. For comparison, we also implemented state-of-the-art self-supervised learning methods (SimCLR~\cite{chen2020simple}, MoCo-v2~\cite{chen2020improved}, BYOL~\cite{grill2020bootstrap}) pretrained on \textbf{Voxceleb1} from scratch, as well as utilizing the model weights pretrained on ImageNet~\cite{deng2009imagenet}, as shown in \tableautorefname~\ref{tab:emo}. In the last four rows of \tableautorefname \ref{tab:emo}, we reported the performance of the facial representation learning by adding each strategy separately or jointly. As noted, our proposed strategies significantly outperform both the ImageNet-pretrained model as well as other self-supervised methods, on all tasks. 

In addition, we compared with one state-of-the-art unsupervised deepfake work, CycleFace~\cite{chang2021learning}, which encompasses an identity-irrelevant branch to encode facial expressions. We finetuned its emotion extraction branch and it yielded inferior performance across all metrics \colorbox{my_color1}{(beige row)}. Our conjecture is that not only expression, but other attributes that are irrelevant to the expression, such as head poses, are encoded to the expression branch as well, which was also raised by the authors \cite{chang2021learning}.

To evaluate the efficacy of \textbf{FaceSwap}, we also trained with CutMix~\cite{yun2019cutmix} as shown in \tableautorefname~\ref{tab:emo} row (c) by only cutting and mixing the centre patch of the images. Comparing it with Ours as shown in row (e), our superior performance demonstrates that the proposed \textbf{FaceSwap} leads to better FER representations.

\vspace{-5pt}\subsection{Qualitative Results}\vspace{-5pt}

We visualised the saliency of contributing features with Grad-CAM~\cite{selvaraju2017grad} across different persons with different emotions. As shown in \figureautorefname~\ref{fig:saliency} (a), the model trained with SimCLR mostly unsatisfactorily attend on emotion-irrelevant areas. By contrast, our proposed \textbf{FaceSwap} and \textbf{MaskFN} focus more on eyes and mouths, indicating that the proposed strategies are able to regulate the network to focus more on the regions that are more expressive to emotions. Moreover, the focus slightly changes with different emotions. ``Neutral'' focuses more on eyes while ``Happy'' and ``Fear'' focus more on eyes and mouths. 
Additionally, we also visualised the saliency maps of the same subject across time in \figureautorefname~\ref{fig:saliency} (b). 

\textcolor{black}{It can be observed that with the same subject, the focus of our method slightly changes as the expression changes, indicating that our network is able to attend on expression-related regions.  }

\input{image/saliency}

\vspace{-12pt}
\subsection{Additional Experiments}
\vspace{-5pt}
\label{subsec:add_exp}

\noindent\textbf{Influence on different strategies:} Table \ref{tab:emo} shows the effectiveness of each module. It can be observed that they are effective for the emotion classification task, while \textbf{MaskFN} leads to poorer result on the Valence \& Arousal Recognition. One possible reason is that our assumption for false negative cancellation is based on concrete categorical emotions, whereas those additional ``positive'' by false negative cancellation may not be that ``positive'' for Valence/Arousal regression. Please see further discussions in Section \ref{sec:discussion}.

\input{table/table_FR}

\noindent\textbf{Facial recognition performance:} \textcolor{black}{We further evaluated our proposed method with face recognition (FR) on the LFW dataset~\cite{huang2008labeled}. We used the pre-trained network to extract the facial representations and performed verification based on KNN, the same settings as CycleFace~\cite{chang2021learning}. As shown in \tableautorefname~\ref{tab:fr}, the performance on FR degraded with our proposed strategy, in line with our objective of removing identity-related information to avoid shortcuts in FER.}

\noindent\textbf{Influence on training stages:} Table \ref{tab:stage} reports the performance of the model on classification when training at different stages, where the second column shows the training epochs and the third column indicates top-1 accuracy of instance discrimination. Self-supervised training might get overfitted after certain stages on facial images, deteriorating the downstreaming FER tasks. \textcolor{black}{This can be observed in both SimCLR and Ours, indicating that the model is likely to learn identity-related information if overtrained.}

\input{table/table_ablation}

\noindent\textbf{Influence on having different number of false negatives:} As shown in \tableautorefname~\ref{tab:ablation}, introducing an additional false negative improves the classification performance by 0.5\% yet impairs the performance on Valence/Arousal regression; this further validating our hypothesis that \textbf{MaskFN} selects false positives that are more ``positive'' in terms of categorical emotions.

\noindent\textbf{Validation of mouth-eye descriptors:} \textcolor{black}{We provide further experiments to validate that the mouth and eye have more emotional-related information and could act as expressive indicators. We used the feature of mouth and eye extracted from fixed ResNet18 (pretrained on ImageNet), which is identical to what we used in MaskFN. Subsequently, linear probing was performed on AffectNet for FER, and KNN on LFW for FR. The result shows that it has superior performance in FER, yet inferior in FR as shown in \tableautorefname~\ref{tab:descriptor}.}

\vspace{-5pt}\subsection{Limitations}\vspace{-5pt}

\label{sec:discussion}

Albeit the improvement \textbf{MaskFN} has brought about to expression classification task, the performance of learned representation on regression task was noticeably decreased. This may probably result from the fact that our \textbf{MaskFN} strategy was built on the assumption of categorical expressions where similar emotions share similar structural characteristics in eyes and mouth. However, this may not apply to the dimensional expression analysis. Same categories may exhibit different ranges of Valence \& Arousal and pushing together these samples may in turn degrade the sensitivity of VA regression. This highlights that a task-specific network design is necessary for reaching a satisfactory downstream result.

%% file: table/table_finetune.tex
\begin{table}[ht]
\centering
    \setlength{\tabcolsep}{3pt}
    \resizebox{0.96\linewidth}{!}{
    \begin{tabular}{@{}llcccccccccccccc@{}}
    \toprule
    \multicolumn{6}{l}{\multirow{3}{*}{\textbf{Pretraining Methods}}} & \multirow{3}{*}{\textbf{Dataset}} & \multicolumn{2}{c}{\textbf{Freeze\textcolor{Line}{-AffectNet}}} & \multicolumn{6}{c}{\textbf{Finetune\textcolor{Line}{-AffectNet}}} & \textbf{Finetune\textcolor{Line}{-FER2013}} \\
    \cmidrule(lr){8-9} \cmidrule(lr){10-15} \cmidrule(lr){16-16}
     & & & & & & & \multicolumn{2}{c}{\textbf{EXPR}} & \multicolumn{2}{c}{\textbf{EXPR}} &  \multicolumn{2}{c}{\textbf{Valence}} &  \multicolumn{2}{c}{\textbf{Arousal}} & \textbf{EXPR} \\ 
    \cmidrule(lr){8-9} \cmidrule(lr){10-11} \cmidrule(lr){12-13} \cmidrule(lr){14-15} \cmidrule(lr){16-16}
    & & & & & & & F1$\uparrow$ & Acc$\uparrow$ & F1$\uparrow$ & Acc$\uparrow$ & CCC$\uparrow$ & RMSE$\downarrow$ & CCC$\uparrow$ & RMSE$\downarrow$ & Acc$\uparrow$ \\
    \midrule
    \multicolumn{6}{l}{Supervised} & ImageNet & - & - & 56.7\% & 56.6\% & 0.563 & 0.462 & 0.480 & 0.376 & 69.12\% \\
    \multicolumn{6}{l}{BYOL~\cite{grill2020bootstrap}}  & VoxCeleb1 & 34.1\% & 37.2\% & 56.3\% & 56.4\% & 0.560 & 0.460 & 0.462 & 0.386 & 68.98\% \\
    \multicolumn{6}{l}{MoCo-v2~\cite{chen2020improved}}  & VoxCeleb1 & 38.0\% & 38.1\% & 56.8\% & 56.8\% & 0.570 & 0.454 & 0.486  & 0.378 & 69.13\% \\
    \multicolumn{6}{l}{SimCLR~\cite{chen2020simple}}  & VoxCeleb1 & 53.4\% & 53.5\% & 57.5\% & 57.7\% & 0.594 & 0.431 & 0.451 & 0.387 & 68.45\% \\
    \rowcolor{my_color1}
    \multicolumn{6}{l}{CycleFace~\cite{chang2021learning}} & VoxCeleb1,2 & - & - & 48.8\% & 49.7\% & 0.534 & 0.492 & 0.436 & 0.383 & 69.86\% \\
    
    \arrayrulecolor{gray}
    \midrule
    \multirow{7}{*}{\rotatebox{90}{Ours}} & & \multicolumn{1}{c}{TimeAug} & HardNeg & FaceSwap & MaskFN \\
    
    & a & \checkmark & & & & VoxCeleb1 & 53.9\% & 54.1\% & 57.8\% & 57.9\% & 0.583 & 0.448 &0.500 & 0.374 & 69.32\% \\
    & b & \checkmark & \checkmark & & & VoxCeleb1 & 55.4\% & 55.5\% & 58.1\% & 58.3\% & 0.594  & 0.437 & 0.500 & 0.373 & 69.60\% \\
    & c & \checkmark & \checkmark & CutMix~\cite{yun2019cutmix} & & VoxCeleb1 & 54.2\% & 54.2\% & 58.3\% & 58.4\% & 0.542 & 0.463 & 0.508 & 0.368 & 69.15\% \\
    & d & \checkmark & & \checkmark & \checkmark & VoxCeleb1 & 55.6\% & 55.8\% & 58.6\% & 58.7\%  & 0.568 & 0.444 & 0.502 & 0.369 & 70.07\% \\
    & e & \checkmark & \checkmark & \checkmark & & VoxCeleb1 & 56.0\% & 56.0\% & 58.8\% & 58.9\% & \textbf{0.601} & \textbf{0.429} & \textbf{0.514}  & \textbf{0.367} & 70.47\% \\
    & f & \checkmark & \checkmark & & \checkmark & VoxCeleb1 & \textbf{57.1\%} & \textbf{57.1\%} & 58.9\% & 58.9\% & 0.578 & 0.448 & 0.493 & 0.370 & 70.21\% \\
    & g & \checkmark & \checkmark & \checkmark & \checkmark & VoxCeleb1 & 56.4\% & 56.4\% & \textbf{59.3\%} & \textbf{59.3\%} & 0.595 & 0.435 & 0.502 & 0.372 & \textbf{71.66\%} \\
    
    \bottomrule
    \end{tabular}}
    \vspace{5pt}
    \caption{\textbf{Comparison of FER performance on AffectNet and FER2013, in terms of categorical expression classification, and valence \& arousal recognition.} The results were based on linear evaluation (Freeze) and fine-tuning (FineTune).}
    \label{tab:emo}
    \vspace{-5pt}
\end{table}

%% file: image/saliency.tex
\begin{figure}[h]
  \vspace{-5pt}
    \centering
    \begin{tabular}{cc}
    \includegraphics[height=0.33\linewidth]{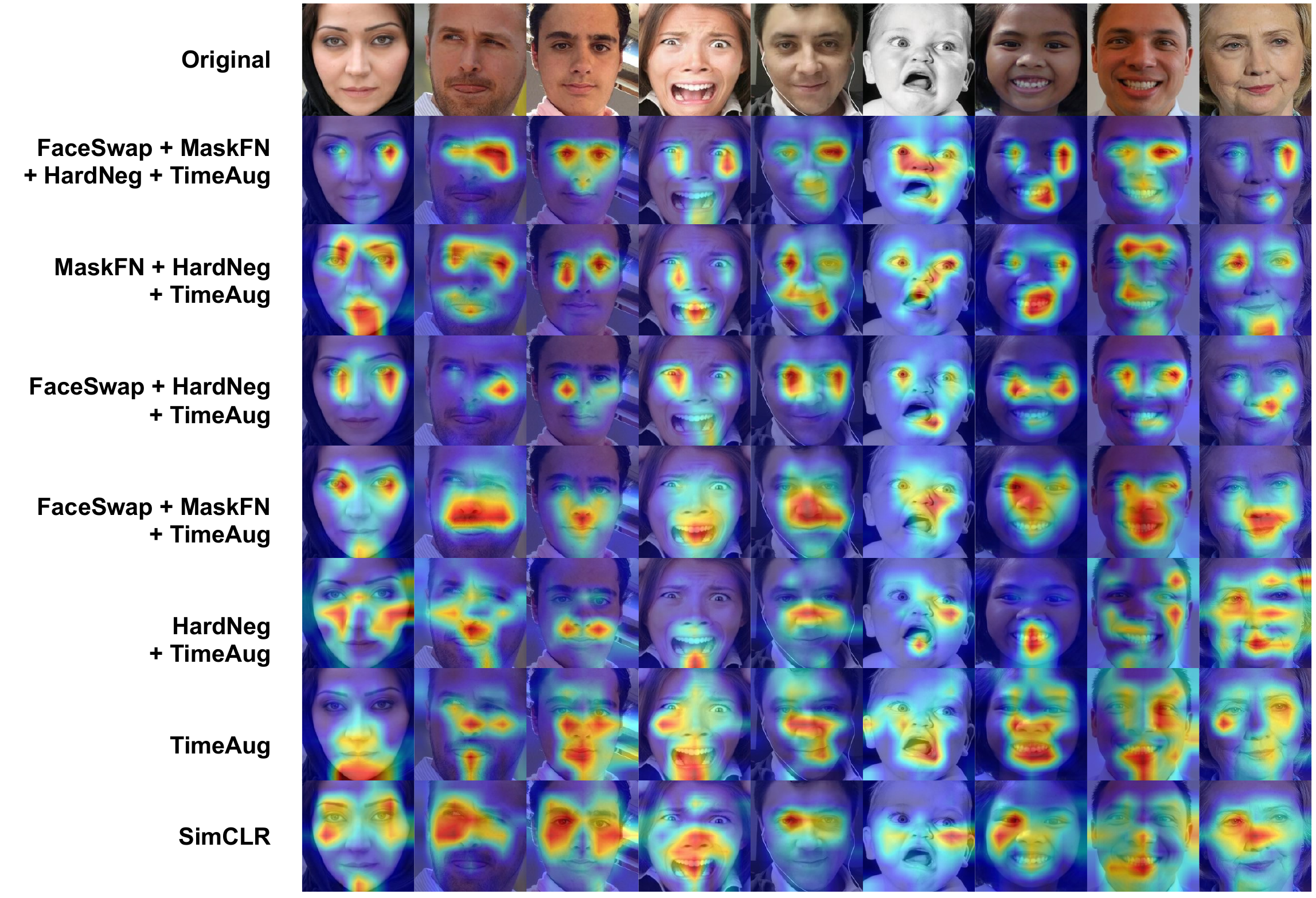}&
    \includegraphics[height=0.33\linewidth]{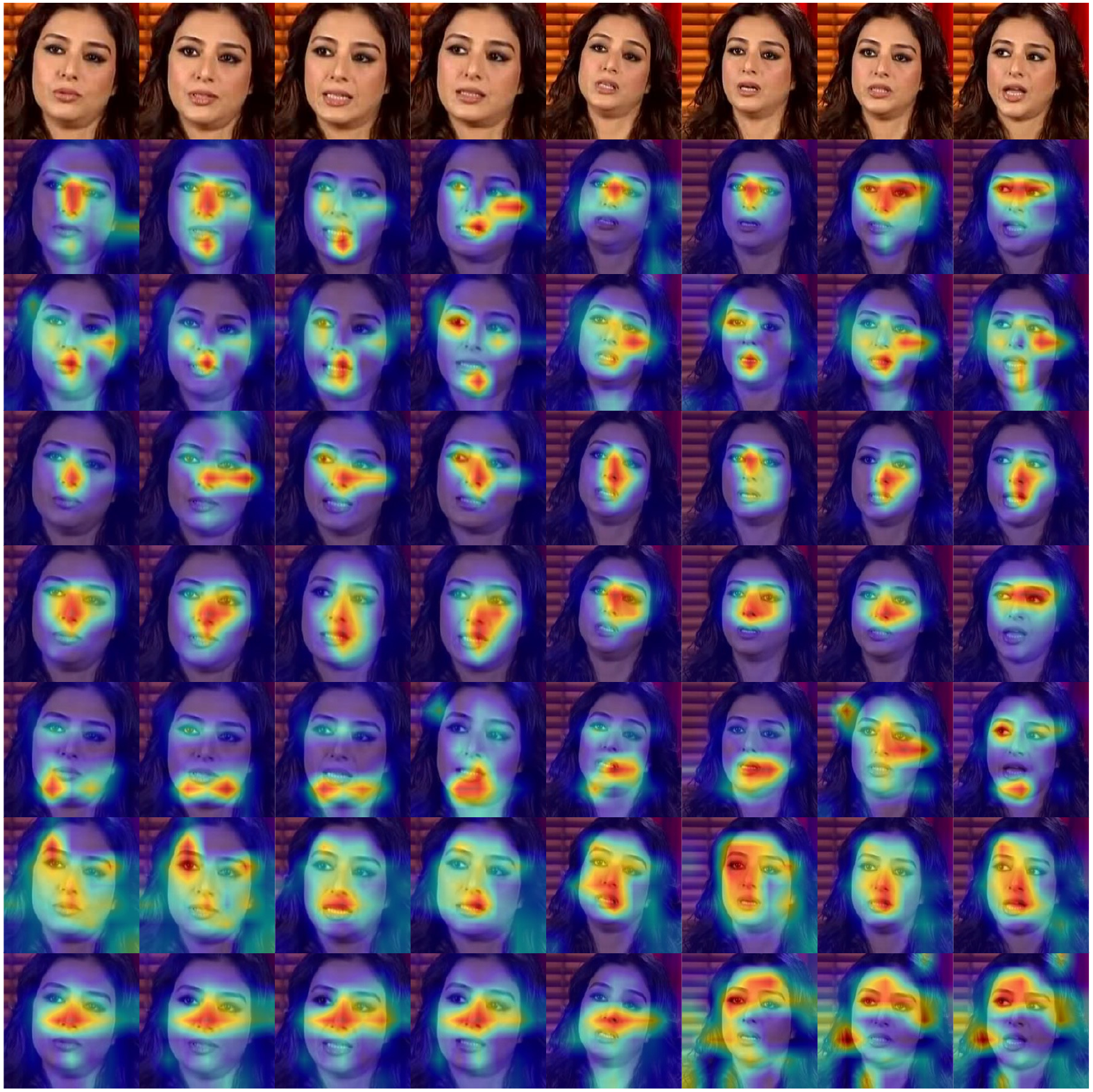}\\
    \small{\; \; \;(a) different persons and emotions} &(b) \small{same person across time}
    \end{tabular}
    \caption{Saliency maps using Grad-CAM~\cite{selvaraju2017grad} comparing different strategies. Pictures are selected from AffectNet~\cite{mollahosseini2017affectnet} (a) and Voxceleb~\cite{nagrani2017voxceleb,chung2018voxceleb2,nagrani2020voxceleb} (b).}
    \vspace{-10pt}
\label{fig:saliency}
\end{figure}

%% file: table/table_FR.tex
\begin{table}[ht]
\centering
  \begin{minipage}[b]{0.48\linewidth}
    \centering
    \setlength{\tabcolsep}{3pt}
    \resizebox{0.9\linewidth}{!}{
    \begin{tabular}{@{}lcccccccc@{}}
    \toprule
    \multicolumn{6}{l}{\textbf{SSL Methods}} & \textbf{FR-KNN} $\downarrow$ \\
    \midrule
    \multicolumn{6}{l}{BYOL~\cite{grill2020bootstrap}}& 64.5\% \\
    \multicolumn{6}{l}{MoCo~\cite{chen2020improved}} & 63.8\% \\
    \multicolumn{6}{l}{SimCLR~\cite{chen2020simple}} & 63.5\% \\
    \arrayrulecolor{gray}
    \midrule
    \multirow{7}{*}{\rotatebox{90}{Ours}} & & \multicolumn{1}{c}{TimeAug} & HardNeg & FaceSwap & MaskFN \\
     & a & \checkmark &  & & & 64.1\% \\
     & b & \checkmark & \checkmark & & & 63.9\% \\
    & c & \checkmark & \checkmark
     & CutMix~\cite{yun2019cutmix} & & 62.9\% \\
     & e & \checkmark &  \checkmark & \checkmark  &  & 59.5\% \\
    & f & \checkmark &  \checkmark &  & \checkmark & 58.2\% \\
    & g & \checkmark &  \checkmark & \checkmark & \checkmark & \textbf{56.5\%} & \\
    \arrayrulecolor{black}
    \bottomrule
    \end{tabular}}
    \vspace{10pt}
    \caption{Face recognition (FR) performance on the LFW dataset using KNN.}
    \label{tab:fr}
  \end{minipage}\hfill
  \begin{minipage}[b]{0.50\linewidth}
    \begin{center}

    \setlength{\tabcolsep}{3pt}
    \resizebox{0.86\linewidth}{!}{
    \begin{tabular}{@{}lccccc@{}}
    \toprule
    \multicolumn{1}{l}{\multirow{2}{*}{\textbf{Methods}}} &\multicolumn{1}{c}{\multirow{2}{*}{\textbf{Epochs}}} & \multirow{2}{*}{\textbf{Top1-Acc}} & \multirow{2}{*}{\textbf{FR-KNN}} & \multicolumn{2}{c}{\multirow{1}{*}\small{\textbf{FER-Finetune}}} \\
    \cmidrule(lr){5-6} 
    & & & & F1$\uparrow$ & Acc$\uparrow$
     \\\midrule
    \multirow{4}{*}{SimCLR} & 5 & 85.7\% & 55.2\% & 52.5\% & 52.7\% \\
    & 10 & 90.8\% & 61.7\% & 55.0\% & 55.1\% \\
     & 30 & 95.3\% & 62.6\% & 56.8\% & 56.9\% \\
     & 50 & 97.1\% & 61.6\% & 57.6\% & 57.7\%  \\
     & 150 & 99.0\% & 61.4\% & 55.4\% & 55.5\% \\
    \arrayrulecolor{gray}
    
    \midrule 
    \multirow{4}{*}{Ours} & 150 & 65.4\% & 59.5\% & 59.1\% & 59.1\% \\
     & 180 & 75.6\% & 58.0\% & 59.2\% & 59.3\% \\
     & 210 & 85.1\% & 56.5\% & 59.3\% & 59.3\% \\
     & 250 & 90.0\% & 58.7\% & 58.3\% & 58.5\% \\
    
    \arrayrulecolor{black}
    \bottomrule
    \end{tabular}}
    
    \end{center}
    \caption{Results of different training stages.} \label{tab:stage}
  \end{minipage}
\vspace{-5pt}
\end{table}

%% file: table/table_ablation.tex
\begin{table}[ht]
\centering
  \begin{minipage}[b]{0.68\linewidth}
    \centering
    \setlength{\tabcolsep}{3pt}
    \resizebox{0.9\linewidth}{!}{
    \begin{tabular}{@{}lcccccc@{}}
    \toprule
    \multirow{2}{*}{\textbf{Methods}} & \multicolumn{2}{c}{\textbf{EXPR}} &  \multicolumn{2}{c}{\textbf{Valence}} &  \multicolumn{2}{c}{\textbf{Arousal}}  \\ \cmidrule(lr){2-3} \cmidrule(lr){4-5} \cmidrule(lr){6-7} 
     & $F_1\uparrow$ & Acc$\uparrow$ & CCC$\uparrow$ & RMSE$\downarrow$ & CCC$\uparrow$ & RMSE$\downarrow$  \\
    \midrule
    MaskFN(1) & 59.3\% & 59.3\% & \textbf{0.595}  & \textbf{0.435} & \textbf{0.502} & \textbf{0.372}  \\
    MaskFN(2)  & \textbf{59.7\%} & \textbf{59.7\%} & 0.583 & 0.444 & 0.484 & 0.378  \\
    \bottomrule
    \end{tabular}}
    \vspace{5pt}
    \caption{Influence on having different numbers (as in brackets) of false negatives. (With TimeAug, HardNeg and FaceSwap)}
    \label{tab:ablation}
  \end{minipage}\hfill
  \begin{minipage}[b]{0.28\linewidth}
    \begin{center}
    \setlength{\tabcolsep}{3pt}
    \resizebox{0.8\linewidth}{!}{
    \begin{tabular}{@{}ccc@{}}
    \toprule
     \multicolumn{1}{c}{\multirow{2}{*}{\textbf{FR-KNN}}} & \multicolumn{2}{c}{\textbf{FER-Linear}} \\\cmidrule(lr){2-3} 
     & F1 & Acc \\
    \midrule
     45.5\% & 51.0\% & 51.1\%  \\ 
    \bottomrule
    \end{tabular}}
    \vspace{15pt}
    \caption{\textbf{Eye-Mouth Descriptor.}}
    \label{tab:descriptor}

    \end{center}
  \end{minipage}

\end{table}

%% file: section/discussion.tex
Facial representations are complicated, composed of multiple attributes. Conventional contrastive based self-supervised learning fails to learn expression-specific representations. In this paper, we revisited the use of self-supervised contrastive learning, and proposed three complementary novel strategies to regulate the network to lean towards emotion related information. In particular, based on emotion related properties in facial images, we explored effective augmentations, hard negative pair sampling manner, as well as false negative cancellation strategy. The experimental results have shown that our self-supervised training strategies outperform the state-of-the-art methods on downstream FER tasks, including both categorical expression classification and dimensional valence\&arousal regression.

%% file: section/appendix.tex
\vspace{15pt}
\section*{Supplementary Materials}
\section{Details of Experiment Setup}
\subsection{Self-supervised pre-training}
\label{subsec:ssl_pre}

\noindent\textbf{Hyper-parameters.}
All the self-supervised learning methods applied ResNet 50 as the backbone, followed by two MLP layers that project the embedding feature into a 128-dimension space. The batch size was set as 256. We optimized the network by Adam with the weight decay as 1e-4 and learning rate initialized as 3e-4. We applied cosine annealing learning rate scheduler from 10 epochs onwards. Empirically, MoCo-v2 and BYOL were optimized by SGD instead.

\noindent\textbf{Data augmentation.} Details of the data augmentation strategies in our proposed contrastive learning are shown in Algorithm~\ref{alg:ssl_aug}. Below are the explanations of the augmentation operations we applied. 

\begin{itemize}
    \item \textbf{TimeAug}: Randomly sample along time domain with $t_1$ time interval, which follows a downscale distribution over $[0, T_1]$. In our case, $T_1=1$ second.
    
    \item \textbf{FaceSwap}: Randomly select an image from the training set and set it as the sample that provides identity information $S_{id}$. Apply FaceSwap to swap facial expression of x. The probability of performing FaceSwap $p_s$ follows a Bernoulli distribution with probability as 0.5.  
    \item \textbf{Mask}: Randomly apply mask on the area of eyes and mouth with the normalised value. Probability $p_m$ follows a Bernoulli distribution (0.8).
    
    \item  \textbf{Resize}: Resize the image to a size of 128 $\times$ 128.
    
    \item \textbf{Crop}: Randomly crop a region from the image with a size of 112 $\times$ 112.
    
    \item \textbf{Flip}: Horizontal flip, with a probability $p_f$ of 0.5
    
    \item \textbf{Jitter}: Colour jittering (0.4 brightness, 0.4 contrast, 0.4 saturation and 0.2 hue). Probability $p_j$ is set to 0.8
    
    \item \textbf{Blur}: Gaussian blur, with a probability $p_b$ under 0.5 Bernoulli distribution.
    
    \item \textbf{Gray}: Grayscale, with a probability $p_g$ of 0.5 Bernoulli distribution.
\end{itemize}

\begin{algorithm}
\begin{footnotesize}
\caption{Data augmentation for self-supervised pre-training.}\label{alg:ssl_aug}
\begin{algorithmic}
\State \textbf{Input}: Images $\boldsymbol{X}=\{\bx_{1}, \bx_{2}, ... \bx_{N}\}$ from video clips

\For{$\bx \in \boldsymbol{X}$}
    \If{$\bx$ is positive}
        \State x'=TimeAug(x, $t_i$) where $t_1 \in T_1$
        \State x'=FaceSwap(x') if $p_s$
    \ElsIf{$\bx$ is anchor}
        \State x'=x
    \EndIf
    \State x'=Mask(x') if $p_m$
    \State x'=Crop(Resize(x'))
    \State x'=Flip(x') if $p_f$
    \State x'=Jitter(x') if $p_j$
    \State x'=Blur(x') if $p_b$
    \State x'=Gray(x') if $p_g$
\EndFor
\State \textbf{Output}: Augmented images $\boldsymbol{X'}=\{\bx'_{1}, \bx'_{2}, ... \bx'_{N}\}$

\end{algorithmic}
\end{footnotesize}
\end{algorithm}

\subsection{Downstream task} 
\vspace{5pt}

\subsubsection{Facial expression recognition}

\noindent\textbf{Hyper-parameters.} We provide results of both freezing (freezing the pre-trained model layers) and fine-tuning (tuning all layers). All downstream tasks of FER (Emotion Classification and Valence \& Arousal Recognition) were trained with a batch size of 64. The network was trained with an Adam optimizer with the weight decay as 5e-4 over 20 epochs for AffectNet dataset and 300 epochs for FER2013 dataset. Initially, the learning rate was set to 1e-4 and decay with cosine annealing learning rate scheduler.

\vspace{5pt}

\begin{algorithm}
\begin{footnotesize}
\caption{Data augmentation for downstream task.}\label{alg:downstream_aug}
\begin{algorithmic}
\State \textbf{Input}: Images $\boldsymbol{X}=\{\bx_{1}, \bx_{2}, ... \bx_{N}\}$ from video clips
 
\For{$\bx \in \boldsymbol{X}$}
    \State x'=Mask(x) if $p_m$
    \State x'=Crop(Resize(x'))
    \State x'=Flip(x') if $p_f$
    \State x'=Jitter(x') if $p_j$
    \State x'=Blur(x') if $p_b$
    \State x'=Gray(x') if $p_g$
\EndFor
\State \textbf{Output}: Augmented images $\boldsymbol{X'}=\{\bx'_{1}, \bx'_{2}, ... \bx'_{N}\}$ 
\end{algorithmic}
\end{footnotesize}
\end{algorithm}

\vspace{10pt}

It should be noted we evaluated the pretrained model of CycleFace~\cite{chang2021learning} with a different set of hyper-parameters,  empirically, on the emotion classification task of FER2013. We optimized the cross-entropy loss using SGD with Nesterov momentum, using a batch size of 64, a weight decay of 1e-4 and a momentum of 0.9. The learning rate was decayed using Reduce learning rate on Plateau scheduler with the initial learning rate of 1e-2.

\noindent\textbf{Data augmentation.} The applied data augmentation of downstream task training, is illustrated in Algorithm~\ref{alg:downstream_aug}. It should be noted that when fine-tuning with CycleFace~\cite{chang2021learning}, we resized the images to (80, 80) and then cropped to (64, 64) because the network only accepts this size of image input.

\subsubsection{Face recognition} We also tested our results on face recognition with KNN to further validate that our method is more robust against other facial attributes, such as face identity. The distance between features is calculated using $L2$ norm, with the same setting as Cycleface~\cite{chang2021learning}.

\section{Additional Results}
\subsection{Computational cost} 

\begin{table}[htp]
    \centering

    \setlength{\tabcolsep}{3pt}
    \resizebox{0.6\linewidth}{!}{
    \begin{tabular}{@{}lccc@{}}
    \toprule
    \textbf{Methods} & \textbf{batchsize} & \textbf{time/150-epochs} & \textbf{memory/GPU} \\
    \midrule
    SimCLR~\cite{chen2020simple} & 256 & 8h & 1.6GB \\
    \arrayrulecolor{gray}
    
    \midrule 
    Ours (All Strategies) & 256 & 17.5h & 2.2GB \\
    
    \arrayrulecolor{black}
    \bottomrule
    \end{tabular}}
    \vspace{10pt}
    \caption{Comparison of time and memory cost between SimCLR and Ours (with Nvidia RTX 3090). }
    \label{tab:cost}
\end{table}

We present the computational cost of our proposed method in~\tableautorefname~\ref{tab:cost}. This was measured under the same experimental settings (e.g., hardware and batchsize). It can be observed that under the same epoch number, it would take more time and more GPU memory for our proposed method. Moreover, as shown in~\tableautorefname~\ref{tab:stage} in the main paper, SimCLR tends to learn short-cuts when the epoch number is still small yet the Top1-instance classification is high. With our proposed series of strategies, although the computation cost is larger, our proposed method can effectively avoid shortcuts such as face identity.

On the other hand, we argue that the proposed FaceSwap is an effective strategy to avoid identity-related shortcuts, by generating intermediate fake faces during the training stages. This simple operation may introduce some artifacts (as shown in~\figureautorefname~\ref{fig:faceswap} of the main paper) in the face-swapped images, compared to those state-of-the-art Deepfake algorithms~\cite{rossler2019faceforensics++,perov2020deepfacelab}. However, it is much more computationally efficient for online data augmentation. 

\subsection{Batchsize sensitivity}

\begin{table}[ht]
    \centering

    \setlength{\tabcolsep}{3pt}
    \resizebox{0.56\linewidth}{!}{
    \begin{tabular}{@{}lccc@{}}
    \toprule
    \multicolumn{1}{l}{\multirow{2}{*}{\textbf{Methods}}} &\multicolumn{1}{c}{\multirow{2}{*}{\textbf{batchsize}}} & \multicolumn{2}{c}{\multirow{1}{*}\small{\textbf{FER-Finetune}}} \\
    \cmidrule(lr){3-4} 
    & & F1$\uparrow$ & Acc$\uparrow$
     \\\midrule
    Ours (TimeAug+HardNeg+MaskFN) & 64 & 56.4\% & 56.5\% \\
    \arrayrulecolor{gray}
    
    \midrule 
    Ours (TimeAug+HardNeg+MaskFN) & 256 & 58.9\% & 58.9\% \\
    
    \arrayrulecolor{black}
    \bottomrule
    \end{tabular}}
    \vspace{10pt}
    \caption{Pre-trained model with different batchsize.}
    \label{tab:batchsize}
\end{table}

We present the results of applying different batchsize during pretraining. As shown in table~\ref{tab:batchsize}, reducing the batchsize would impair the performance. We think this issue could come from two perspectives:

\begin{itemize}
    \item Contrastive learning itself is sensitive to batchsize since the core InfoNCE loss applied in contrastive learning has been proven to benefit from large batch sizes \cite{chen2020simple}. 
    \item Our method for False Negative Cancellation was designed based on the categorical expression assumption, as discussed in the Section~\ref{subsec:add_exp} of the main paper. That is, when the batch size is much larger than the downstream category number, the facial images with similar mouth-eye descriptors have higher chances of being the same category. Therefore, a larger batchsize would increase the probability of correctly picking up false-negative samples.
\end{itemize}